%% file: main.tex
\documentclass[10pt,twocolumn,letterpaper]{article}

\usepackage[pagenumbers]{cvpr} 

\input{preamble}
\definecolor{cvprblue}{rgb}{0.21,0.49,0.74}
\usepackage[pagebackref,breaklinks,colorlinks,allcolors=cvprblue]{hyperref}
\usepackage{algorithmic}
\usepackage{algorithm}
\usepackage{graphicx}
\usepackage{multirow}
\usepackage{arydshln}
\usepackage{wrapfig}

\def\paperID{*****} 
\def\confName{CVPR}
\def\confYear{2026}

\title{Coded-E2LF: Coded Aperture Light Field Imaging from Events}

\author{Tomoya Tsuchida$^\dagger$\hspace{5mm}
Keita Takahashi$^\dagger$\hspace{5mm}
Chihiro Tsutake$^\dagger$\hspace{5mm}
Toshiaki Fujii$^\dagger$\hspace{5mm}
Hajime Nagahara$^\ddagger$ \\
$^\dagger$ Nagoya University, Japan\hspace{5mm}
$^\ddagger$ Osaka University, Japan
}

\begin{document}
\maketitle
\input{sec/00_abst}    
\input{sec/01_intro}
\input{sec/02_related}
\input{sec/03_method}

\input{sec/04_exp}
\input{sec/05_conc}

{
    \small
    \bibliographystyle{ieeenat_fullname}
    \bibliography{refs}
}

\input{sec/07_supple}

\end{document}

%% file: sec/00_abst.tex
\begin{abstract}

We propose Coded-E2LF (coded event to light field), a computational imaging method for acquiring a 4-D light field using a coded aperture and a stationary event-only camera. In a previous work, an imaging system similar to ours was adopted, but both events and intensity images were captured and used for light field reconstruction. In contrast, our method is purely event-based, which relaxes restrictions for hardware implementation. We also introduce several advancements from the previous work that enable us to theoretically support and practically improve light field reconstruction from events alone. In particular, we clarify the key role of a black pattern in aperture coding patterns. We finally implemented our method on real imaging hardware to demonstrate its effectiveness in capturing real 3-D scenes. To the best of our knowledge, we are the first to demonstrate that a 4-D light field with pixel-level accuracy can be reconstructed from events alone. Our software and supplementary video are
available from our project website: \href{https://www.fujii.nuee.nagoya-u.ac.jp/Research/EventLF/}{https://www.fujii.nuee.nagoya-u.ac.jp/Research/EventLF/}.

\end{abstract}

%% file: sec/01_intro.tex
\section{Introduction}

The notion of a light field enables us to extend the representation of visual information from the conventional 2-D spatial domain to the 4-D spatial-angular domain. The application of light fields includes depth estimation~\cite{honauer2017dataset,shin18epinet,khan2021edgeaware}, object/material recognition~\cite{Maeno2013,wang20164d}, view synthesis~\cite{Kalantari2016,mildenhall2019llff, broxton2020immersive}, and 3-D display~\cite{wetzstein2012tensor,huang2015light,lee2016additive}. In this paper, we consider how to acquire a light field efficiently.

A light field usually consists of a large number of images arranged in a 2-D grid with small viewpoint intervals (e.g., $8 \times 8$ views). A brute-force approach for light-field imaging is to directly sample all pixels at all viewpoints~\cite{wilburn2005high,fujii2006multipoint,Taguchi2009}. As a more efficient alternative, coded-aperture imaging~\cite{liang2008programmable,nagahara2010programmable,Inagaki_2018_ECCV,Sakai_2020_ECCV,Guo2022TPAMI} has also been developed, in which light-attenuating coding patterns are placed in the aperture plane of a camera to optically encode the incoming light rays before they reach the image sensor. With this method, only a small number (e.g., 2--4) of images taken from a stationary camera with different coding patterns are sufficient to reconstruct a light field~\cite{Inagaki_2018_ECCV,Guo2022TPAMI,Tateishi_2022_IEICE}. 

To enhance the efficiency of coded-aperture light-field imaging, Habuchi et al.~\cite{Habuchi_2024_CVPR} used an event camera, which can detect intensity changes as events asynchronously at each pixel. In their method, intensity changes caused by the change in coding patterns were detected as events and used for light-field reconstruction. However, an intensity image, captured simultaneously with the events, was also used. Therefore, a hybrid camera was required that can take both events and intensity images. Due to the dependence on intensity images, their method went only halfway in exploiting the potential of event-based sensing. 

We propose a purely event-based method for coded-aperture light field imaging, called Coded-E2LF, which stands for \textit{coded event to light field}. In contrast to Habuchi et al.'s~\cite{Habuchi_2024_CVPR}, our method eliminates the need for intensity images, which relaxes restrictions for hardware implementation. Our method is constructed on Habuchi et al.'s~\cite{Habuchi_2024_CVPR}; we used their algorithm pipeline (with small modifications) as the baseline. However, we introduce several advancements from previous work that enables us to theoretically support and practically improve light-field reconstruction from events alone; in particular, we clarify the key role of a black pattern in the aperture coding patterns. We theoretically analyze the event generation process with a coded aperture and derive two key properties: (i) approximate equivalence between event-based and intensity-based coded aperture imaging in the presence of a black pattern and (ii) approximate permutation invariance of coding patterns. Moreover, on the basis of the theoretical analysis, we introduce two improvements to the baseline algorithm: (i) a black-first coding sequence (BF for short) and (ii) reference-aware event generation (RA for short). We finally implemented our method on real imaging hardware to demonstrate its effectiveness in capturing real 3-D scenes. Although our method was designed under the static-scene assumption, it was applicable to slowly moving scenes thanks to the moderate measurement time ($\simeq$ 30 ms).

To the best of our knowledge, we are the first to demonstrate that a 4-D light field with pixel-level accuracy can be reconstructed from events alone. Our study will contribute to both the progress of light-field imaging and the exploration of the full potential of event-based sensing. 

%% file: sec/02_related.tex
\section{Related Works}

\subsection{Event-based Sensing}

Event cameras~\cite{eventcamerasurvey2022,Brandli_2014_IEEE} feature bio-inspired sensors that can detect intensity changes as events asynchronously at each pixel. Compared with conventional intensity-based sensors, event sensors excel in temporal resolution, dynamic range, and data efficiency. Therefore, event cameras find many applications such as optical-flow estimation, camera-pose estimation, slow-motion video synthesis, and 3-D reconstruction. Some of the methods for these applications require hybrid cameras that can take both intensity images and events, but others work with events only. 

 From the perspective of hardware implementation, event-only methods are more advantageous than those relying on hybrid (intensity + event) inputs. Off-the-shelf hybrid sensors are very limited in performance; e.g., an iniVation DAVIS346 has only 346$\times$260 pixels and a 12M events/sec throughput. A far better performance is available with an event-only sensor; e.g., a Prophesee EVK4 features 1280$\times$720 pixels and 1.066G events/sec~\cite{finateu2020}. A hybrid camera can also be implemented by combining an event-only sensor and an intensity-based sensor via a beam splitter~\cite{messikommer2022multi,Tulyakov21cvpr}, but it inevitably involves additional hardware/software complexities.

Intensity reconstruction from events is a challenging issue. Event-to-video (E2VID) methods~\cite{Rebecq19e2v,scheerlinck2020fast,stoffregen2020eccv,cadena2021spade,paredes2021back,Weng_2021_ICCV,ercan2024hypere2vid} were developed to reconstruct a video (a sequence of intensity images) from events. However, either the target scene or the camera needs to move so as to trigger the events necessary for the reconstruction. Moreover, in general cases, reconstruction of pixel-level intensities from events alone is an ill-posed problem; therefore, E2VID methods were usually designed to pursue perceptual quality and temporal consistency rather than pixel-level accuracy. EventNeRF~\cite{rudnev2023eventnerf} reconstructed a 3-D object from events alone. However, the target object needs to be mounted on a rotating stage in front of a constant color background. Bao et al.~\cite{Bao2025evtemmap} reconstructed an intensity
image from the events that were captured with a gradually opening aperture, while Han et al.~\cite{han2023high} used events recorded during the turn-on process of a light for intensity reconstruction. Although both Bao et al.'s and Han et al.'s methods worked for static scenes with a stationary camera, they were limited to 2-D images. 

In contrast to the previous methods, our method can reconstruct a 4-D light field with \textit{pixel-level accuracy} from events alone. Moreover, our method works for static scenes with a stationary camera because the events necessary for reconstruction are caused actively by the coded aperture.

\subsection{Light-field Imaging}

We first mention light-field imaging methods using conventional intensity-based cameras. The most straight-forward approach to light-field imaging is to use an array of cameras~\cite{wilburn2005high,fujii2006multipoint,Taguchi2009}, which involves costly and bulky hardware. Alternatively, placing a lens array before the image sensor enables light-field imaging in a single shot~\cite{adelson1992single,arai1998gradient,ng2005light,ng2006digital}. However, this method has a trade-off between the number of views and the spatial resolution of each view. Another approach is to place light-attenuating coding patterns before the image sensor~\cite{veeraraghavan2007dappled,liang2008programmable,nagahara2010programmable,marwah2013compressive,Inagaki_2018_ECCV,nabati2018colorLF,vadathya2019,Tateishi_2021_ICIP,vargas_2021_ICCV,Guo2022TPAMI,Mizuno_2022_CVPR,Tateishi_2022_IEICE}. Specifically, with the coded-aperture method, only two to four images, captured from a stationary camera with different coding patterns, are sufficient to reconstruct a light field with $8 \times 8$ views at the full-sensor resolution~\cite{Inagaki_2018_ECCV,Guo2022TPAMI,Tateishi_2022_IEICE}. 

Some early attempts to use event cameras for light-field imaging have been reported. Habuchi et al.~\cite{Habuchi_2024_CVPR} combined a coded aperture with an event camera. However, their method depended on intensity images as well as events. In contrast, we eliminate the need for intensity images, which relaxes the restriction for hardware implementation. In other works~\cite{qu2024eventfields,guo2024eventlfm}, event cameras were combined with a lens array, kaleidoscope, and galvanometer-based optics for light-field imaging, but their focus was on the optical designs rather than the methods/theories for light-field reconstruction. In contrast, we, for the first time, demonstrate that a 4-D light field with pixel-level accuracy can be reconstructed from events alone.

Compared with intensity-based coded aperture imaging, our method has the potential to reduce the measurement time for light fields even in low light conditions. A short measurement time is favorable for capturing dynamic scenes, because events induced by a coded aperture can dominate those generated by scene motion. Meanwhile, low-light conditions are challenging for high-speed intensity-based imaging, especially when using a coded aperture that further attenuates the incoming light.  

\subsection{Computational Imaging and Deep Optics}

Various computational imaging methods have been developed on the basis of deep optics~\cite{Iliadis_2016_mask,Nie_2018_CVPR, Yoshida_2018_ECCV,Inagaki_2018_ECCV,Wu_2019_ICCP,Li_2020_ICCP,Mizuno_2022_CVPR,Wang_2023_ICCV}, with which the camera-side optical elements and computational algorithms are jointly optimized in a deep-learning-based framework. Our method follows the concept of deep optics to fully exploit the potential of a coded aperture and an event camera with the power of data-driven knowledge. Moreover, along with some recent works~\cite{takatani2021event,han2023high,Habuchi_2024_CVPR,Bao2025evtemmap,Yu_2024_eventps}, our method can be regarded as one of the early attempts to apply computational imaging methods to event cameras, which opens a new horizon for event-based sensing.

%% file: sec/03_method.tex
\section{Proposed Method}
\label{sec:prop}

\begin{figure*}[ht]
\centering
\includegraphics[width=0.9\linewidth]{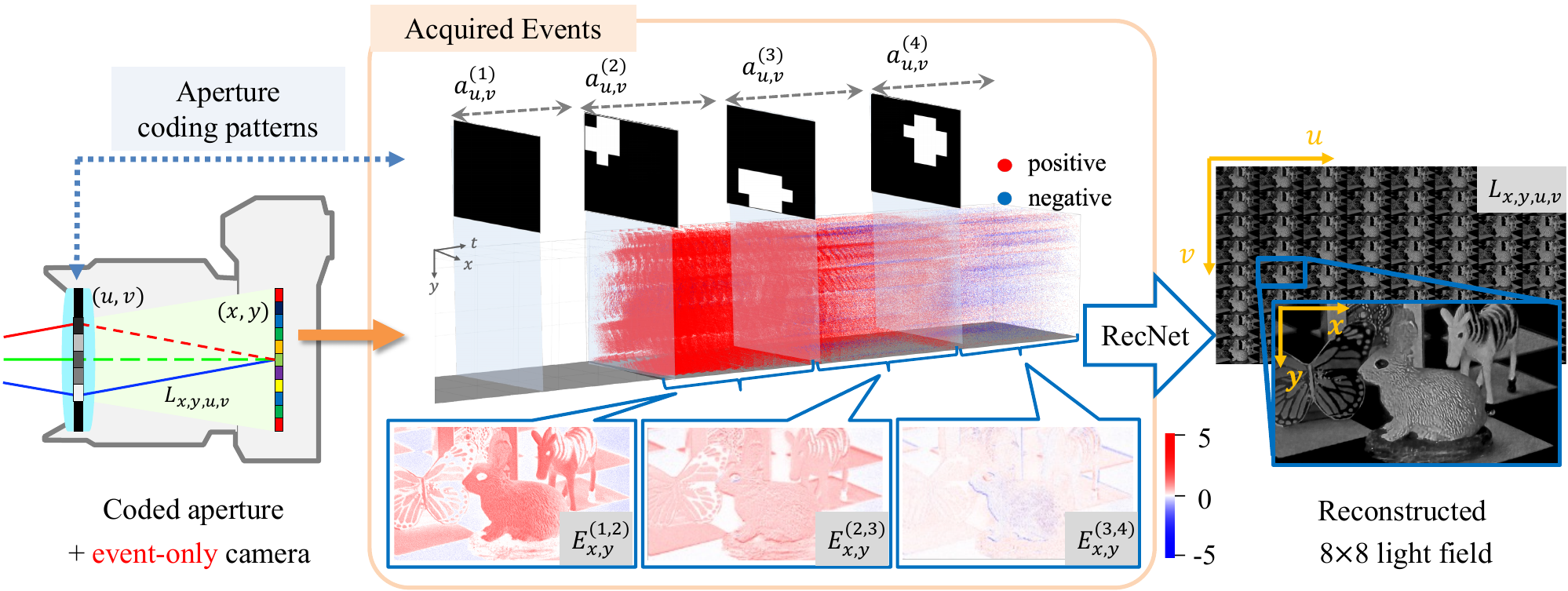}
\caption{Overview. Sequence of coding patterns is applied to aperture plane to generate events. Events are accumulated into event images and then used to reconstruct light field through learned reconstruction algorithm (RecNet). Coding patterns are optimized with RecNet.}
\label{fig:overview}
\end{figure*}

\subsection{Background and Basics}

We define a light field $L$ in the 4-D space $(x,y,u,v)$ to parameterize all the light rays that come into a camera; $(u,v)$ and $(x,y)$ denote the positions on the aperture and imaging planes, respectively (see Fig.~\ref{fig:overview}). We assume that $x,y,u,v$ take discretized integer values; thus, $L$ is equivalent to a set of multiview images in which $(x,y)$ and $(u,v)$, respectively, correspond to the pixel and viewpoint. We assume $L$ has $W  \times  H$ pixels and $8 \times 8$ viewpoints; $L \in \mathcal{R}^{W \times H \times 8 \times 8}$. Each element of $L$ is described using subscripts, i.e., $L_{x,y,u,v}$. We also assume that $L_{x,y,u,v}$ has a monochrome intensity in the range of $[0,1]$. 

Our method aims to acquire a 4-D light field $L$ from a stationary event-only camera as shown in Fig.~\ref{fig:overview}. Before describing our method, we summarize the previous methods for coded aperture imaging using a conventional intensity-based camera and a hybrid event camera.

\textbf{Intensity-based coded aperture imaging}~\cite{liang2008programmable,nagahara2010programmable,Inagaki_2018_ECCV,Sakai_2020_ECCV,Guo2022TPAMI}. 
 A sequence of coding patterns, $a^{(1)}, \dots, a^{(N)}$ , where $a^{(n)}_{u,v} \in [0,1]$ ($u,v \in \{1,\dots,8\}$), is placed at the aperture plane. Let $I^{(n)}$ be the intensity image formed under the $n$-th coding pattern $a^{(n)}$, which is described as
\begin{align}
I^{(n)}_{x,y} = \textstyle\sum_{u,v} a^{(n)}_{u,v}L_{x,y,u,v}.
\label{eq:coded-aperture}
\end{align}
$N$ coded intensity images, $I^{(1)}, \dots, I^{(N)}$ ($I^{(n)} \in \mathcal{R}^{W \times H}$), captured under the $N$ coding patterns are used for light field reconstruction. With the power of deep learning, a relatively small $N$ (e.g. $N=4$) is sufficient for accurate light-field reconstruction~\cite{Inagaki_2018_ECCV,Guo2022TPAMI,Tateishi_2022_IEICE}.

\textbf{Coded aperture + hybrid event camera}~\cite{Habuchi_2024_CVPR}. To enhance the efficiency of coded-aperture light-field imaging, Habuchi et al.~\cite{Habuchi_2024_CVPR} introduced an event camera that can obtain both intensity images and events simultaneously. In their method, $N$ coding patterns ($a^{(1)}, \dots, a^{(N)}$) are applied in sequence \textit{within} a single exposure of an intensity image. Consequently, $I^{(1)}, \dots, I^{(N)}$ cannot be individually recorded, but their sum, given as $ {I}^{\textrm{sum}} = \textstyle\sum_{n=1}^{N} I^{(n)}$, is recorded as a single intensity image. At the same time, events are recorded at each pixel asynchronously during the exposure. They assume that the target scene does not move; thus, these events are caused exclusively by the change of coding patterns. An event is described as $e_{x,y,t}\in \{+1, -1\}$, where $t$ is the time stamp, and the positive/negative signs correspond to the increase/decrease in the intensity at pixel $(x,y)$. During the change from $a^{(n-1)}$ to $a^{(n)}$, more than one event with some delay can be recorded at each pixel; the number of events at $(x,y)$ is roughly proportional to the logarithmic difference between $I^{(n-1)}_{x,y}$ and $I^{(n)}_{x,y}$. Therefore, all the relevant events are accumulated to an event image, $E^{(n-1,n)} \in \mathcal{Z}^{W \times H}$, as
\begin{align}
E^{(n-1,n)}_{x,y} =\textstyle\sum_{t\: \in T^{(n-1,n)}} e_{x,y,t},
\label{eq:EventStack}
\end{align}
where $T^{(n-1,n)}$ denotes the transient time between $a^{(n-1)}$ and $a^{(n)}$. In summary, a single intensity image (${I}^{\textrm{sum}}$) and $N-1$ event images ($E^{(1,2)},\dots, E^{(N-1,N)}$) are captured from the target scene with a single exposure and used for light-field reconstruction.

\textbf{Our method}. The main limitation of Habuchi et al.'s method~\cite{Habuchi_2024_CVPR} is the dependence on a hybrid (intensity + event) camera. In contrast, our method is purely event-based, aiming to reconstruct the light field $L$ only from $N-1$ event images, $E^{(1,2)},\dots, E^{(N-1,N)}$.

\subsection{Theoretical Analysis}
\label{subsec:theory}

Since we are the first to establish a purely event-based method for coded-aperture light-field imaging, we present a theoretical analysis showing why a light field can be reconstructed from events \textit{alone}. Specifically, we analyze the equivalence between our method and traditional intensity-based coded-aperture imaging. Habuchi et al.~\cite{Habuchi_2024_CVPR} presented a similar theoretical analysis to ours, but their analysis was based on the assumption that an intensity image was also obtained. We also analyze the permutation of coding patterns, which helps our algorithm design in Sec.~\ref{sec:black-first}.

According to the mechanism of an event sensor, an event is triggered at pixel $(x,y)$ when
\begin{eqnarray}
|\ln(I_{x,y} + \epsilon) - \ln(I_{x,y}^{\mathrm{ref}} + \epsilon)| > \tau
\label{eq:event-model}
\end{eqnarray}
is satisfied. $I_{x,y}$ is the present intensity at $(x,y)$, $I_{x,y}^{\mathrm{ref}}$ is the reference intensity when the last event was recorded at $(x,y)$, $\epsilon$ is a small bias corresponding to the dark current, and $\tau$ is a contrast threshold. 

We assume that a sequence of aperture coding patterns $\{a^{(1)}, \dots, a^{(N)}\}$ is applied to an event camera, and event images, $\{ E^{(1,2)}, \dots, E^{(N-1,N)} \}$, are observed. The corresponding coded intensity images, $\{ I^{(1)}, \dots, I^{(N)}\}$, are unknown, but we can establish approximate relations between the intensity images and event images. On the basis of Eq.~(\ref{eq:event-model}), we can relate an event image ($E^{(n-1,n)}$) with two consecutive intensity images ($I^{(n)}$ and $I^{(n-1)}$) as
\begin{eqnarray}
\tau E^{(n-1, n)}_{x,y} \approx \ln\left(I^{(n)}_{x,y} + \epsilon \right) - \ln\left(I^{(n-1)}_{x,y} + \epsilon \right),
\label{eq:relation}
\end{eqnarray}
where $I_{x,y}^{\mathrm{ref}}$ before the pattern transition is approximated as $I^{(n-1)}_{x,y}$. Equation~(\ref{eq:relation}) includes uncertainties due to the quantization step $\tau$, the ambiguity of $I_{x,y}^{\mathrm{ref}}$, and sensor noise. We can extend Eq.~(\ref{eq:relation}) to the relation between \textit{any} two intensity images ($I^{(n^{\prime})}$ and $I^{(n^{\prime\prime})}$ with $n^{\prime} < n^{\prime\prime}$) as
\begin{eqnarray}
 \sum_{n=n^{\prime}+1}^{n^{\prime\prime}} \tau E^{(n-1,n)}_{x,y} \approx \ln\left(I^{(n^{\prime\prime})}_{x,y} + \epsilon \right) - \ln\left(I^{(n^{\prime})}_{x,y} + \epsilon \right). \!
\label{eq:relation2} \!\!
\end{eqnarray}

\textbf{Approximate equivalence between event-based and intensity-based coded-aperture imaging}. Reconstruction of $N$ intensity images ($I^{(1)}, \dots, I^{(N)}$) from $N-1$ event images ($E^{(1,2)}, \dots, E^{(N-1,N)}$) is an ill-posed problem because the number of unknowns ($N$) is greater than that of knowns ($N-1$). However, there is a special case where all the intensity images can be reconstructed. Suppose that one of the aperture coding patterns is set to black; $a^{(n)}_{u,v} = 0$ for all $(u,v)$s with $n=n_{\textrm{B}}$. In this case, the corresponding intensity image is also black; $I^{(n_{\textrm{B}})}_{x,y} = 0$ for all $(x,y)$s, and the other $N-1$ intensity images ($I^{(n)}$ with $n \neq n_{\textrm{B}}$) can be obtained from Eq.~(\ref{eq:relation2}) as 
\begin{eqnarray}
I^{(n)}_{x,y}  \approx \left\{ \!\!\!
\begin{array}{ll}
\epsilon \left\{\exp\left(\tau \!\!\!\!\!\displaystyle\sum_{k=n_{\textrm{B}}+1}^{n} \!\!E^{(k-1,k)}_{x,y} \! \right) \! - 1 \right\} &\!\!\!\! (n > n_{\textrm{B}}) \\
\epsilon \left\{\exp\left(- \tau \!\!\!\!\!\displaystyle\sum_{k=n+1}^{n_{\textrm{B}}} \!\!E^{(k-1,k)}_{x,y} \!\right) \! - 1 \right\} &\!\!\!\! (n < n_{\textrm{B}}) 
\end{array}
\right. \!\!\!
\label{eq:relation3}.\!\!\!\!
\end{eqnarray}
To conclude, in the presence of a black pattern, event-based coded aperture imaging is equivalent to intensity-based coded aperture imaging because $\{ E^{(1,2)},\dots,E^{(N-1,N)}\}$ and $\{ I^{(1)},\dots,I^{(N)}\}$ (with $I^{(n_{\textrm{B}})}_{x,y} = 0$) are interconvertible via Eq.~(\ref{eq:relation3}). Note that this equivalence is not strict due to the uncertainties of Eq.~(\ref{eq:relation}); related to this point, $I^{(n)}$s obtained by Eq.~(\ref{eq:relation3}) are harshly quantized by the contrast threshold $\tau$ (since the value of $E^{(n-1,n)}_{x,y}$ is limited to integers). 

We found that the coding patterns learned with the baseline algorithm (introduced in Sec.~\ref{subsec:baseline}) always include at least one black pattern; see Sec.~\ref{sec:supple1} for examples. The theory presented here is a possible explanation for this result; machine learning chooses the solution that ensures equivalence to intensity-based coded-aperture imaging. In contrast, the coding patterns learned with Habuchi et al.~\cite{Habuchi_2024_CVPR} did not include a black pattern; the black pattern was unnecessary when an intensity image $I^{\textrm{sum}}$ was also obtained.

\textbf{Approximate permutation invariance of coding patterns}. The sequence of coding patterns used to record $E^{(1,2)}, \dots, E^{(N-1,N)}$ is referred to as the original one. We hypothesize a case where the order of the coding patterns is permuted arbitrarily. By using Eqs.~(\ref{eq:relation}) and (\ref{eq:relation2}), we can generate a \textit{virtual} event image, which would be produced if a transition from $I^{(n^{\prime})}$ to $I^{(n^{\prime\prime})}$ ($n^{\prime}$ and $n^{\prime\prime}$ are arbitrary numbers) occurs, as
\begin{eqnarray}
\tilde{E}^{(n^{\prime},n^{\prime\prime})}_{x,y}  \approx \left\{ \!\!\!
\begin{array}{ll}
\textstyle\sum_{n=n^{\prime}+1}^{n^{\prime\prime}} E^{(n-1,n)}_{x,y} &\!\!\!\! (n^{\prime} < n^{\prime\prime}) \\
\textstyle\sum_{n=n^{\prime\prime}+1}^{n^{\prime}} -E^{(n-1,n)}_{x,y} &\!\!\!\! (n^{\prime\prime} < n^{\prime})
\end{array}
\right. 
\label{eq:virtual-event}.\!\!\!\!
\end{eqnarray}
With this equation, any virtual event image that would be produced with the permuted sequence can be computed from $E^{(1,2)}, \dots, E^{(N-1,N)}$. Therefore, the coding patterns are permutation invariant in the sense that the event images are interconvertible between the original and permuted pattern sequences via Eq.~(\ref{eq:virtual-event}). Note that this invariance is not strict due to the uncertainties of Eq.~(\ref{eq:relation}).


\subsection{Baseline Algorithm}
\label{subsec:baseline}

Habuchi et al.'s method~\cite{Habuchi_2024_CVPR} was designed to use both intensity and event images. We found that their method, in its original format, does not work well in the absence of intensity images (see Sec.~\ref{sec:exp}). Therefore, we introduce a baseline algorithm that is technically equivalent to Habuchi et al.'s~\cite{Habuchi_2024_CVPR} but is slightly modified from the original. The changes are summarized as follows; (i) we discard the intensity image $I^{\textrm{sum}}$ from the observed data, (ii) we modify the noise and hyper-parameters in the event generation process to make them better suit the camera we use (a Prophesee EVK4), and (iii) we remove the complementary constraint for the coding patterns to allow more freedom for them. Further improvements over the baseline algorithm will be mentioned in Sec.~\ref{sec:black-first}.

\textbf{Overview}. The baseline algorithm consists of two trainable modules, AcqNet and RecNet. AcqNet describes the data acquisition process using a coded aperture and an event camera, generating event images from an original light field. The trainable parameters in AcqNet correspond to the $N$ coding patterns for the aperture plane. Meanwhile, RecNet receives the event images as input and aims to reconstruct the light field. The AcqNet-RecNet pipeline is trained to minimize the mean-squared error between the original and reconstructed light fields. Once the training is complete, AcqNet is replaced with real imaging hardware, in which the coding patterns are adjusted to the learned parameters of AcqNet. The event data acquired from the camera are fed to RecNet to reconstruct  a real 3-D scene.

\begin{algorithm}[tb]
    \caption{Pseudo-code for AcqNet}
    \begin{algorithmic}[1]     
    \STATE $\dot{a}^{(n)} \in {\cal R}^{8 \times 8} $  \hspace{4mm}\# trainable tensors for coding pattern.
    \STATE forward($L$, $s$): \hspace{2mm} \# $L$: input light field, $s$: scale param. 
    \STATE \hspace{\algorithmicindent} \textbf{for} $n$ in [$1, \dots, N$]:
    \STATE \hspace{2\algorithmicindent} $a^{(n)} = \mbox{sigmoid}(s \dot{a}^{(n)})$
    \STATE \hspace{2\algorithmicindent} compute $I^{(n)}$ by Eq.~(\ref{eq:coded-aperture}) 
    \STATE \hspace{\algorithmicindent} \textbf{end}
    \STATE \hspace{\algorithmicindent} \textbf{for} $n$ in [$2, \dots, N$]:
    \STATE \hspace{2\algorithmicindent} compute $E^{(n-1,n)}$ by Eq.~(\ref{eq:ESIM})  
    \STATE \hspace{\algorithmicindent} \textbf{end}    
    
    \STATE \hspace{\algorithmicindent} return $E^{(1,2)}$, $\dots$, $E^{(N-1,N)}$
    \end{algorithmic}
\end{algorithm}

\textbf{AcqNet}. Algorithm 1 describes a pseudo-code of AcqNet, which takes a light field $L$ as input and produces a sequence of event images $\{ E^{(1,2)}, \dots, E^{(N-1,N)} \}$. The implementation of AcqNet basically follows Habuchi et al.'s~\cite{Habuchi_2024_CVPR}. We assume that the coding patterns should take binary (0/1) values. To generate the coding patterns, we use $N$ sets of trainable tensors, each with $8\times8$ elements, denoted as $\dot{a}^{(n)}$ ($n\in \{1,\dots,N\}$). They are multiplied by the scale parameter $s$ and then fed to the sigmoid function to produce the coding patterns $a^{(n)}$. As the training proceeds, $s$ gradually increases, which forces $a^{(n)}$ to gradually converge to binary patterns. The event generation process in Line 8 is formulated in accordance with Eq.~(\ref{eq:event-model}) as
\begin{align}
\!\!\! E^{(n-1,n)}_{x,y} \!\!= Q\!\left(\! \dfrac{\ln(\bar{I}^{(n)}_{x,y} + \epsilon) - \ln(\bar{I}^{(n-1)}_{x,y} + \epsilon) + w_{x,y}}{\tau + z_{x,y}}\!\!\right)\!\!
\label{eq:ESIM} 
\end{align}
where $Q(x) = \mbox{sign}(x)\mbox{floor}(|x|)$ is a quantization operator,\footnote{A {gradient-pass-through} technique is used to make it differentiable.} and $\bar{I}^{(n)}_{x,y}={I}^{(n)}_{x,y}/64$ is the intensity at pixel $(x,y)$ under the $n$-th coding pattern normalized by the number of views. We use zero-mean Gaussian noises, $w_{x,y}$ and $z_{x,y}$ ($\sigma$ was set to 0.175 and 0.04, respectively), to account for the randomness of the sensor, and $\epsilon=0.01$ and $\tau=0.30$ as default parameters. These parameters were experimentally chosen to suit the camera we used (a Prophesee EVK4).

\textbf{RecNet}. The input to RecNet is a sequence of event images, which are stacked along the channel dimension to form an $(N-1) \times H \times W $ tensor. The output from RecNet is a $64 \times H \times W$ tensor, where 64 channels correspond to 64 (8 $\times$ 8) views of the reconstructed light field. Since architecture design is not the focus of this paper, we use the same architecture (a CNN with 23 layers) as Habuchi et al.~\cite{Habuchi_2024_CVPR} for a fair comparison; see Sec.~\ref{sec:supple-recnet}.

\subsection{Black-first Coding Sequence}
\label{sec:black-first}

In this subsection, we introduce two improvements to the baseline algorithm that can reduce the number of events while improving the quality of light fields.

\textbf{Black-first coding sequence (BF)}. As shown in Fig.~\ref{fig:apertures}, a black pattern is always included in the sequence of coding patterns learned with the baseline algorithm. However, the position of the black pattern is random. We also observe that the pattern transitions before and after the black pattern cause significant numbers of events. Therefore, placing the black pattern in the middle of the sequence is inefficient in terms of the total event count. If the black pattern is located at the start of the sequence, the total event count can be drastically reduced. According to the permutation invariance in Sec.~\ref{subsec:theory}, the position of the black pattern would not significantly affect the information included in the event images. For these reasons, we use a black-first coding sequence (BF for short), in which $a^{(1)}$ is fixed to the black pattern but the remaining $N-1$ coding patterns ($a^{(n)}$ with $n\ge 2$) are learned through AcqNet-RecNet training. As shown in Fig.~\ref{fig:apertures}, BF is effective to reduce the total event count.

Reducing the event count leads to a shorter measurement time in theory. With a sufficiently short measurement time, we can capture a moving scene while ignoring scene motion during the measurement. The theoretical throughput of an EVK4 camera is 1.066G events/sec with 1280$\times$720 pixels~\cite{finateu2020}, which corresponds to 1.16K events/sec per pixel. Therefore, if we need to acquire 7.18 events per pixel (the average event count with the full implementation of our method), the theoretical lower-bound for the measurement time is 6.2 ms.  With a real coded aperture camera, the measurement time cannot be shortened to the theoretical limit due to delay and hardware restrictions. However, pushing down the number of events is beneficial for stable hardware realization and would be more so with better hardware implementations in the future. 

\begin{figure}[t]
\centering
\includegraphics[width=\linewidth]{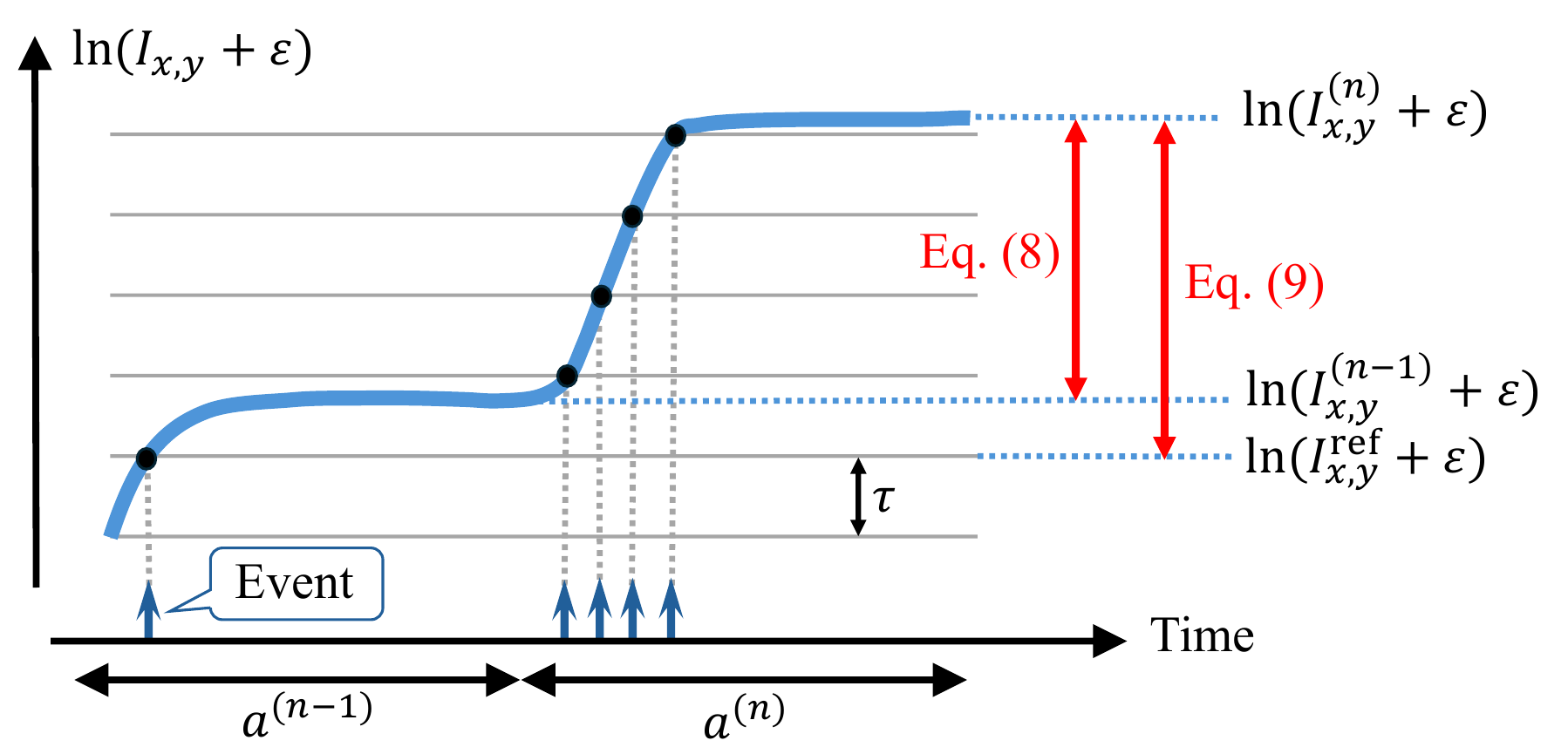}
\caption{Reference-aware event generation: Eq.~(9) uses  $I^{\mathrm{ref}}_{x,y}$ to better simulate event generation process than Eq.~(8).}
\label{fig:RA}
\end{figure}

\textbf{Reference-aware event generation (RA)}. The event generation process in the baseline algorithm (Eq.~(\ref{eq:ESIM})) has a significant limitation; $E^{(n-1,n)}_{x,y}$ is computed from $I^{(n)}_{x,y}$ and $I^{(n-1)}_{x,y}$ without directly using the reference intensity. The same was true with Habuchi et al.'s method~\cite{Habuchi_2024_CVPR}; see Eq.~(12) in their paper. To strictly follow Eq.~(\ref{eq:event-model}), $E^{(n-1,n)}_{x,y}$ should be computed from $I^{(n)}_{x,y}$ and $I^{\mathrm{ref}}_{x,y}$, in which $I^{\mathrm{ref}}_{x,y}$ is the reference intensity just before the transition from $a^{(n-1)}$ to $a^{(n)}$. This reference-aware event generation (RA) is hard to implement because $I^{\mathrm{ref}}_{x,y}$ is uncertain in general cases. However, it can be implemented with a black-first coding sequence; we initialize $I^{\mathrm{ref}}_{x,y} = 0$ for all $(x,y)$s at $n=1$ and use Eqs.~(\ref{eq:ESIM2}) and (\ref{eq:ESIM2-update}) instead of Eq.~(\ref{eq:ESIM}).
\begin{align}
&\!\! E^{(n-1,n)}_{x,y} = Q\!\left( \dfrac{\ln(\bar{I}^{(n)}_{x,y} + \epsilon) - \ln(I^\mathrm{ref}_{x,y} + \epsilon) + w_{x,y}}{\tau + z_{x,y}}\right) \label{eq:ESIM2} \\
& \ln(I^{\mathrm{ref}}_{x,y} + \epsilon ) \gets \ln(I^{\mathrm{ref}}_{x,y} + \epsilon)  + \tau E^{(n-1,n)}_{x,y}
\label{eq:ESIM2-update}.
\end{align}
In the above, we first compute $E^{(n-1,n)}_{x,y}$ from $I^{(n)}_{x,y}$ and $I^{\mathrm{ref}}_{x,y}$ by Eq.~(\ref{eq:ESIM2}) and then update the term related to $I^{\mathrm{ref}}_{x,y}$ by Eq.~(\ref{eq:ESIM2-update}). As shown in Fig.~\ref{fig:RA}, accounting for $I^{\mathrm{ref}}_{x,y}$ can result in different event counts; three events using Eq.~(\ref{eq:ESIM}) versus four events using Eq.~(\ref{eq:ESIM2}) in this example.

As shown in Sec.~\ref{sec:exp}, simply applying BF to the baseline algorithm slightly degrades the quality of light fields. However, when BF is combined with RA, it provides better light-field quality with fewer events than the baseline algorithm.

%% file: sec/04_exp.tex
\section{Experiments}
\label{sec:exp}

\begin{figure}[t]
\newcommand{\hs}{\hspace{-1mm}}
\newcommand{\myvs}{\vspace{1mm}}
\newcommand{\HS}{\hspace{2mm}}
\newcommand{\BS}{\hspace{-13mm}}
\newcommand{\bs}{\hspace{-11mm}}
\centering
    \small
    \begin{tabular}{c|cccc}
      Methods & \HS $a^{(1)}$ & \HS $a^{(2)}$ & \HS $a^{(3)}$ & \HS $a^{(4)}$ \\
      (total events) &  & \BS $|E^{(1,2)}|$ & \BS $|E^{(2,3)}|$ & \BS $|E^{(3,4)}|$ \\ \hline\\[-3mm]
      \hs\raisebox{5mm}{Habuchi}\hs\hs & 
      \hs\includegraphics[width=12mm,height=12mm]
      {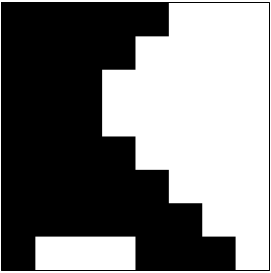}\hs\hs\hs\hs &
      \hs\includegraphics[width=12mm,height=12mm]
      {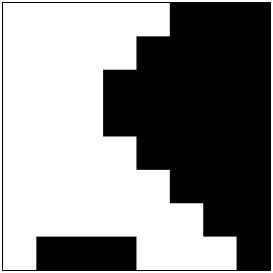}\hs\hs\hs\hs & 
      \hs\includegraphics[width=12mm,height=12mm]
      {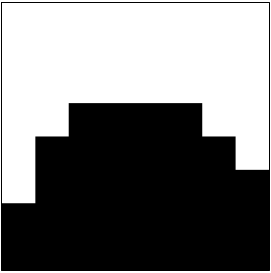}\hs\hs\hs\hs &  
      \hs\includegraphics[width=12mm,height=12mm]
      {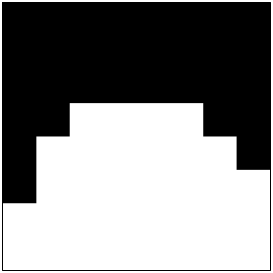}\hs\hs\hs\hs \\
      (0.133) &  & \bs 0.044 & \bs 0.029 & \bs 0.060 \\ \hline \\[-3mm]
      \hs\hs\hs\raisebox{5mm}{Habuchi event-only}\hs\hs\hs & 
      \hs\includegraphics[width=12mm,height=12mm]
      {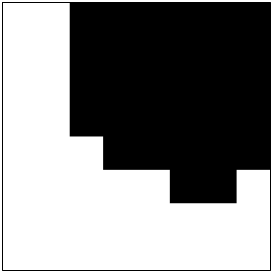}\hs\hs\hs\hs &
      \hs\includegraphics[width=12mm,height=12mm]
      {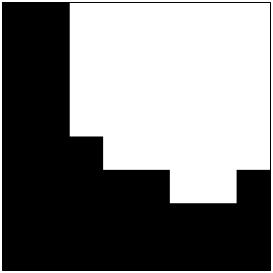}\hs\hs\hs\hs & 
      \hs\includegraphics[width=12mm,height=12mm]
      {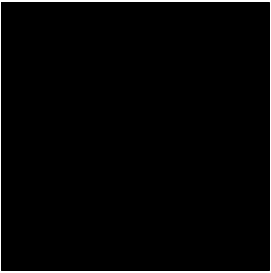}\hs\hs\hs\hs &  
      \hs\includegraphics[width=12mm,height=12mm]
      {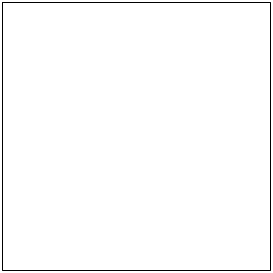}\hs\hs\hs\hs \\
      (21.008) &  & \bs 0.051 & \bs \textbf{9.313} & \bs \textbf{11.644} \\ \hline \\[-3mm]
      \raisebox{5mm}{Baseline} & 
      \hs\includegraphics[width=12mm,height=12mm]{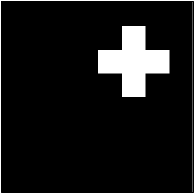}\hs\hs\hs\hs &
      \hs\includegraphics[width=12mm,height=12mm]{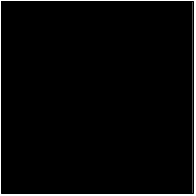}\hs\hs\hs\hs & 
      \hs\includegraphics[width=12mm,height=12mm]{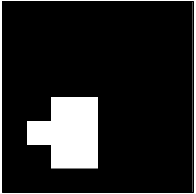}\hs\hs\hs\hs &  
      \hs\includegraphics[width=12mm,height=12mm]{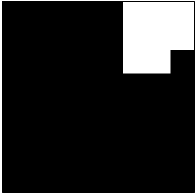}\hs\hs\hs\hs \\
      (9.492) & & \bs \textbf{4.179} & \bs \textbf{5.049} & \bs 0.264 \\ \hline \\[-3mm]  
      \raisebox{5mm}{Baseline+BF} & 
      \hs\includegraphics[width=12mm,height=12mm]{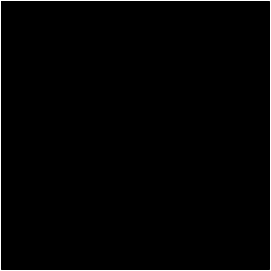}\hs\hs\hs\hs &
      \hs\includegraphics[width=12mm,height=12mm]{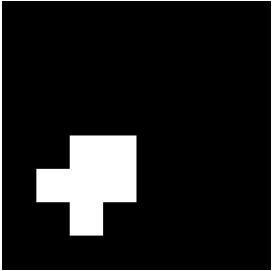}\hs\hs\hs\hs & 
      \hs\includegraphics[width=12mm,height=12mm]{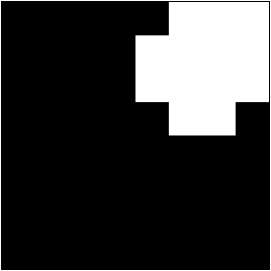}\hs\hs\hs\hs &  
      \hs\includegraphics[width=12mm,height=12mm]{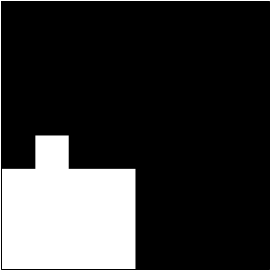}\hs\hs\hs\hs \\
      (6.522) & & \bs \textbf{4.646} & \bs 1.642 & \bs 0.234 \\ \hline \\[-3mm]  
      \raisebox{5mm}{\hs\textbf{Baseline+BF+RA}\hs} & 
      \hs\includegraphics[width=12mm,height=12mm]{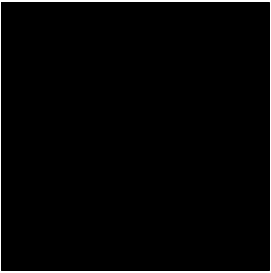}\hs\hs\hs\hs &
      \hs\includegraphics[width=12mm,height=12mm]{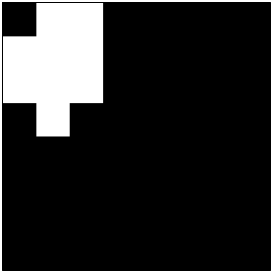}\hs\hs\hs\hs & 
      \hs\includegraphics[width=12mm,height=12mm]{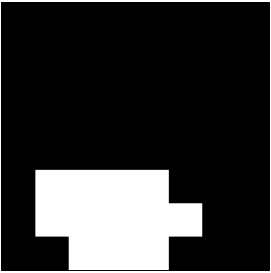}\hs\hs\hs\hs &  
      \hs\includegraphics[width=12mm,height=12mm]{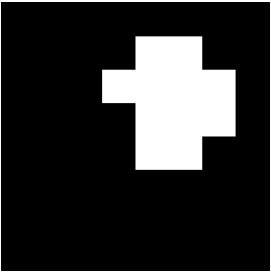}\hs\hs\hs\hs \\
      (7.175) &  & \bs \textbf{5.717} & \bs 1.086 & \bs 0.372        
    \end{tabular}
    \caption{Learned coding patterns ($N=4$) with event counts in $E^{(1,2)}$, $E^{(2,3)}$, and $E^{(3,4)}$ and their total. Event counts are per pixel, averaged in BasicLFSR test dataset. Many events are caused before and after black pattern (numbers in \textbf{bold}).}
    \label{fig:apertures}
\end{figure}

\begin{table*}[t]
\centering
    \caption{Quantitative comparison of our method against previous methods on BasicLFSR test dataset. Reported scores are PSNR[dB] /SSIM; larger is better for both. CA+E2VID requires one-hot coding patterns for 64 individual viewpoints. Habuchi requires additional intensity image as input. Baseline is technically equivalent to Habuchi event-only but with slight modifications. Baseline+BF+RA corresponds to full implementation of our method. Contrast threshold $\tau$ was set to 0.30 for all methods.}
    \small
    \label{tab:quantitative_comparison}
    \begin{tabular}{l|c||ccccc|c}
      Methods & $N$ & EPFL & HCI (new) & HCI (old) & INRIA & Stanford & ALL \\ \hline 
      \hline
      CA+E2VID~\cite{Rebecq19e2v} & 64 &  14.30/0.4537 & 18.97/0.5916 & 19.71/0.6360 & 14.24/0.5590 & 18.05/0.6428  & 17.05/0.5766 \\
      CA+E2VID~\cite{stoffregen2020eccv} & 64 &  14.84/0.4296 & 18.29/0.4814 & 18.53/0.4982 & 14.25/0.4766 & 17.54/0.5689  & 16.69/0.4909 \\ 
      CA+E2VID~\cite{Weng_2021_ICCV} & 64 &  15.32/0.4544 & 18.66/0.5088 & 19.08/0.5706 & 14.49/0.5115 & 18.51/0.5889  & 17.21/0.5268 \\ 
      CA+E2VID~\cite{ercan2024hypere2vid} & 64 &  14.90/0.4115  & 17.29/0.4601 & 18.00/0.5119 & 14.53/0.4634 & 15.83/0.4959  & 16.11/0.4686 \\ \hline
      Habuchi~\cite{Habuchi_2024_CVPR} & 4  & 33.05/0.9341 & 29.62/0.7808 & 35.06/0.8956 & 34.64/0.9364 & 26.81/0.8175  & 31.83/0.8729 \\
      Habuchi event-only~\cite{Habuchi_2024_CVPR} & 4 & 28.18/0.8584 & 26.49/0.6887 & 31.43/0.8085 & 29.45/0.8727 & 23.58/0.7022 & 27.83/0.7861  \\ \hline
      Baseline & 4 & 29.50/0.8669 & 28.64/0.7477 & 33.25/0.8513 & 31.20/0.8804 & 26.45/0.8048  & 29.81/0.8302 \\
      Baseline+BF& 4 &  29.33/0.8636 & 28.53/0.7496 & 32.97/0.8512 & 31.09/0.8818 & 26.53/0.8145  & 29.69/0.8321 \\
      \textbf{Baseline+BF+RA}& 4 &  30.40/0.8835 & 29.06/0.7523 & 34.03/0.8645 & 32.00/0.8909 & 26.92/0.8152  & 30.48/0.8413 \\\hdashline[1pt/1pt]
      \textbf{Baseline+BF+RA}& 8 & 31.78/0.9109 & 29.98/0.7865 & 35.52/0.8999 & 33.38/0.9111 & 27.90/0.8440  & 31.71/0.8705\\
    \end{tabular}
\end{table*}

\begin{figure*}[t]
    \centering
    \newlength{\W}
    \setlength{\W}{33mm}
    {
    \fontsize{9}{9.5}\selectfont
    \tabcolsep = 1pt
    \begin{tabular}{cccccc}
        \includegraphics[height=28mm,keepaspectratio]
        {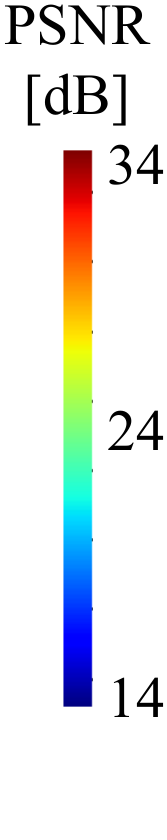} &
        \includegraphics[width=\W,height=\W,keepaspectratio]
        {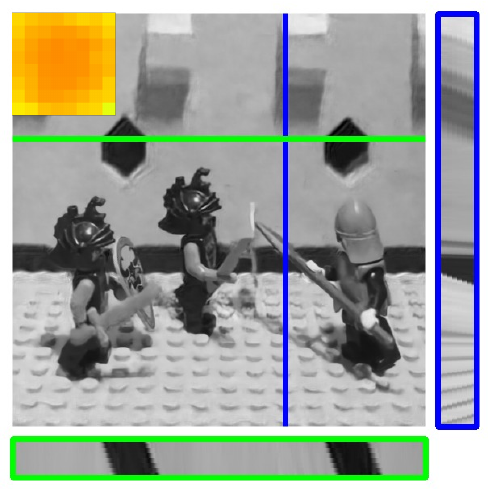} &
        \includegraphics[width=\W,height=\W,keepaspectratio]
        {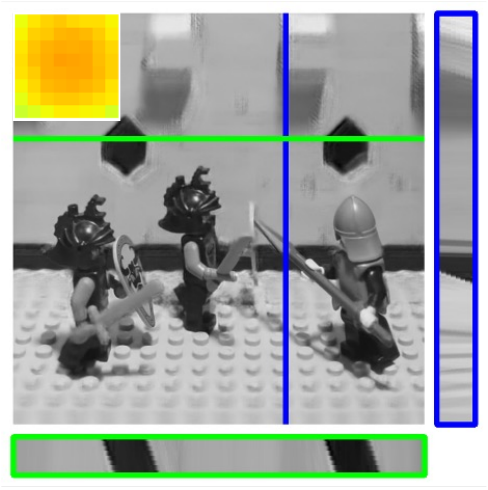} &
        \includegraphics[width=\W,height=\W,keepaspectratio]
        {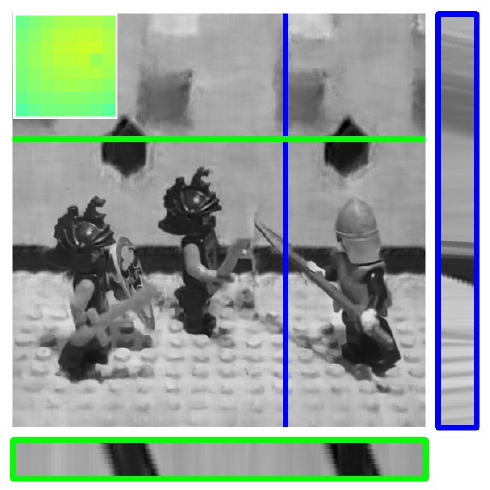} &
        \includegraphics[width=\W,height=\W,keepaspectratio]
        {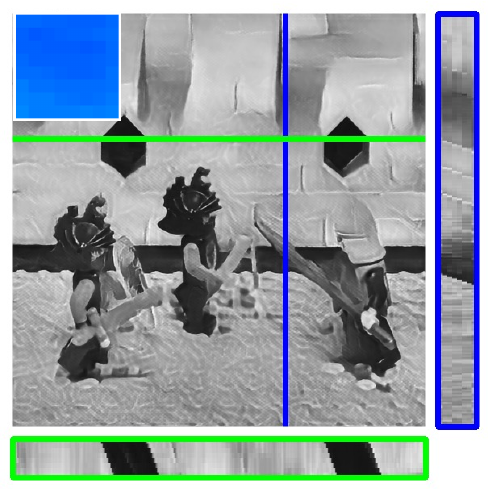} &
        \includegraphics[width=\W,height=\W,keepaspectratio]
        {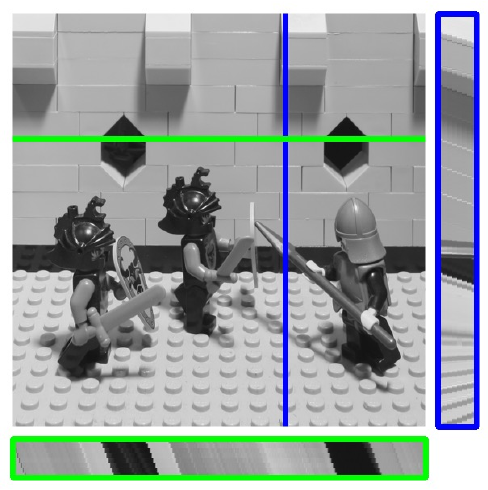}\\[-1mm]
        &
        \textbf{Baseline+BF+RA} ($N$=4) & 
        Habuchi~\cite{Habuchi_2024_CVPR} &
        Habuchi event-only~\cite{Habuchi_2024_CVPR}&
        CA+E2VID~\cite{Weng_2021_ICCV} & 
        Ground truth \\
    \end{tabular}
    }
    \caption{Visual result with \textit{Lego Knights} light field: reconstructed top-left views with epipolar plane images along blue/green lines and insets showing view-by-view reconstruction quality in terms of PSNR.}
    \label{fig:sim-exp}
    \vspace{-1mm}
\end{figure*}

\subsection{Simulation Experiments}

The algorithm pipeline (AcqNet and RecNet) was implemented using PyTorch 2.7.1 and trained on the BasicLFSR dataset~\cite{BasicLFSR}. Following Habuchi et al.~\cite{Habuchi_2024_CVPR}, we extracted 29,327 training samples, each with $64 \times 64$ pixels and $8\times8$ views, from 144 light fields designated for training. We used the built-in Adam optimizer with default parameters and trained the AcqNet-RecNet pipeline over 600 epochs with a batch size of 16. The parameter $s$ in Algorithm 1 was initialized as 1 and multiplied by 1.02 for each epoch.

Simulation experiments were conducted on the BasicLFSR test dataset~\cite{BasicLFSR}, which included 23 light fields categorized into five groups (EPFL, HCI(new), HCI(old), INRIA, and Stanford), all of which were reserved for evaluation. For each light field, we computationally performed data acquisition (AcqNet) and reconstruction (RecNet) and quantitatively evaluated the reconstruction accuracy. 

The learned coding patterns and the event counts per pixel obtained with them (averaged in the BasicLFSR test dataset) are shown in Fig.~\ref{fig:apertures}. Quantitative scores (PSNR/SSIM) of the reconstructed light fields are summarized in Table~\ref{tab:quantitative_comparison}. Some visual results are presented in Fig.~\ref{fig:sim-exp}. See the supplementary video and appendix for more results. 

\textbf{Variants of our method}. We implemented three variants of our method: \textbf{Baseline}, \textbf{Baseline+BF}, and \textbf{Baseline+BF+RA}, where Baseline+BF+RA corresponds to the full implementation of our method. Baseline+BF+RA was implemented with $N=4$ and $N=8$, but we mainly discuss the result with $N=4$. As shown in Fig.~\ref{fig:apertures}, the coding patterns learned with Baseline always included at least one black pattern, but its position was random. By placing the black pattern at the start of the sequence (Baseline+BF and Baseline+BF+RA), we drastically reduced the number of events. As shown in Table~\ref{tab:quantitative_comparison}, Baseline+BF slightly degraded the quality of light fields compared with Baseline. However, Baseline+BF+RA ($N=4$) provided better quality and a lower number of events than Baseline. See Sec.~\ref{sec:supple2} for more results.

\textbf{Comparision with Habuchi et al.~\cite{Habuchi_2024_CVPR}}. As the most relevant work to ours, Habuchi et al.~\cite{Habuchi_2024_CVPR} used both an intensity image and events captured with a coded aperture (\textbf{Habuchi}). They also reported a case where the input was limited to events (\textbf{Habuchi event-only}) as an ablation study. These two variants were tested with $\tau=0.30$ in our environment.\footnote{For Habuchi, we used the model trained and published by Habuchi et al.~\cite{Habuchi_2024_CVPR}. For Habuchi event-only, we retrained a model using the software provided by them.} As shown in Fig.~\ref{fig:apertures}, the coding patterns of Habuchi event-only also included a black pattern and produced a significant number of events (21.008 events per pixel in total). Although Habuchi event-only is technically equivalent to Baseline, the latter provided much fewer events and better quantitative scores than the former (see Fig.~\ref{fig:apertures} and Table~\ref{tab:quantitative_comparison}), mainly due to the removal of the complementary constraint. Baseline+BF+RA ($N=4$) achieved further better performance, a 2.65 dB improvement in quality and 66 \% reduction in total event count compared with Habuchi event-only. Baseline+BF+RA ($N=8$) almost reached the scores of Habuchi, showing the promising potential of our purely event-based approach.

\begin{wrapfigure}{r}[0pt]{.4\linewidth}
\centering
\vspace{-2mm}
\includegraphics[width=.99\linewidth]{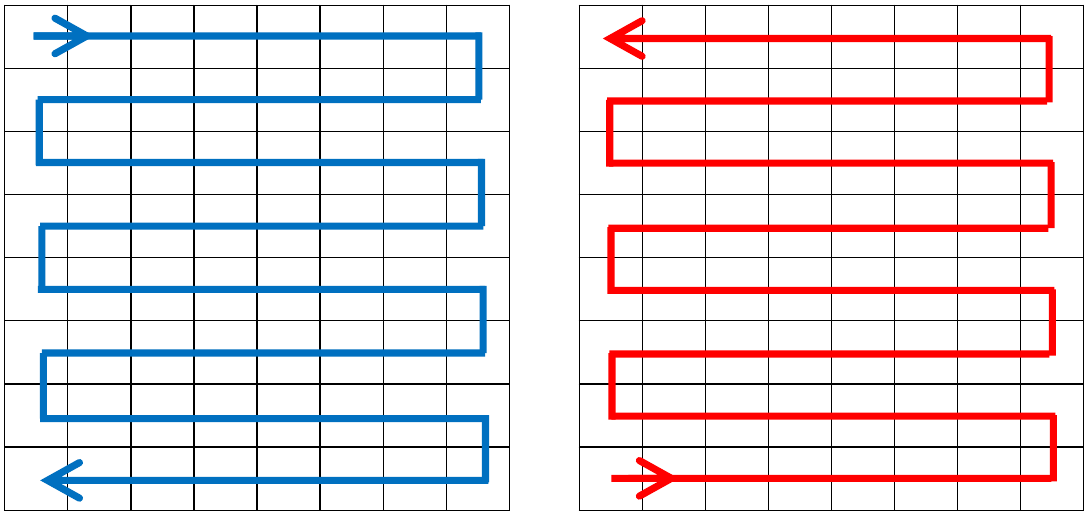}
\caption{Scan order for CA+E2VID: blue/red path for first/second scan.}
\label{fig:scan-orders}
\end{wrapfigure}

\textbf{Comparison with CA+E2VID}.  We also considered a hypothetical scenario in which a coded aperture and E2VID methods were combined for light-field imaging (\textbf{CA+E2VID}). Specifically, a sequence of {one-hot} coding patterns (in which only one viewpoint opened at a time) was used to generate an event stream from the target light field $L$; then, from the event stream, a light field $\hat{L}$ was reconstructed as a video by using an E2VID method.
To simulate this scenario, we created a video with 127 frames that scanned $8 \times 8$ views twice as shown in Fig.~\ref{fig:scan-orders}, and we generated an event stream from the video by using v2e~\cite{hu2021v2e} with $\tau=0.30$. We used the EVREAL software (with pre-trained weights)~\cite{ercan2023evreal} to execute several E2VID methods~\cite{Rebecq19e2v,stoffregen2020eccv,Weng_2021_ICCV,ercan2024hypere2vid}. We used the frames reconstructed from the second scan as $\hat{L}$, which were better than those from the first scan. 
$\hat{L}$ was scale-corrected for quantitative evaluation; the scale was found by $\arg\min_{\eta} \| L - \eta \hat{L}\|^2$.
As shown in Table~\ref{tab:quantitative_comparison} and Fig.~\ref{fig:sim-exp}, the results obtained with CA+E2VID may seem plausible, but they are significantly inaccurate in terms of pixel-level intensities. In contrast, our method can accurately reconstruct the pixel-level intensities.

\begin{figure*}
\centering
\setlength{\tabcolsep}{1.0pt}
\newlength{\apertureW}
\newlength{\eventW}
\setlength{\apertureW}{17.5mm}
\setlength{\eventW}{30mm}

\newlength{\cameraW}
\newlength{\lfW}
\setlength{\cameraW}{103mm}
\setlength{\lfW}{70mm}
\begin{tabular}{cc}
\includegraphics[width=\cameraW]
{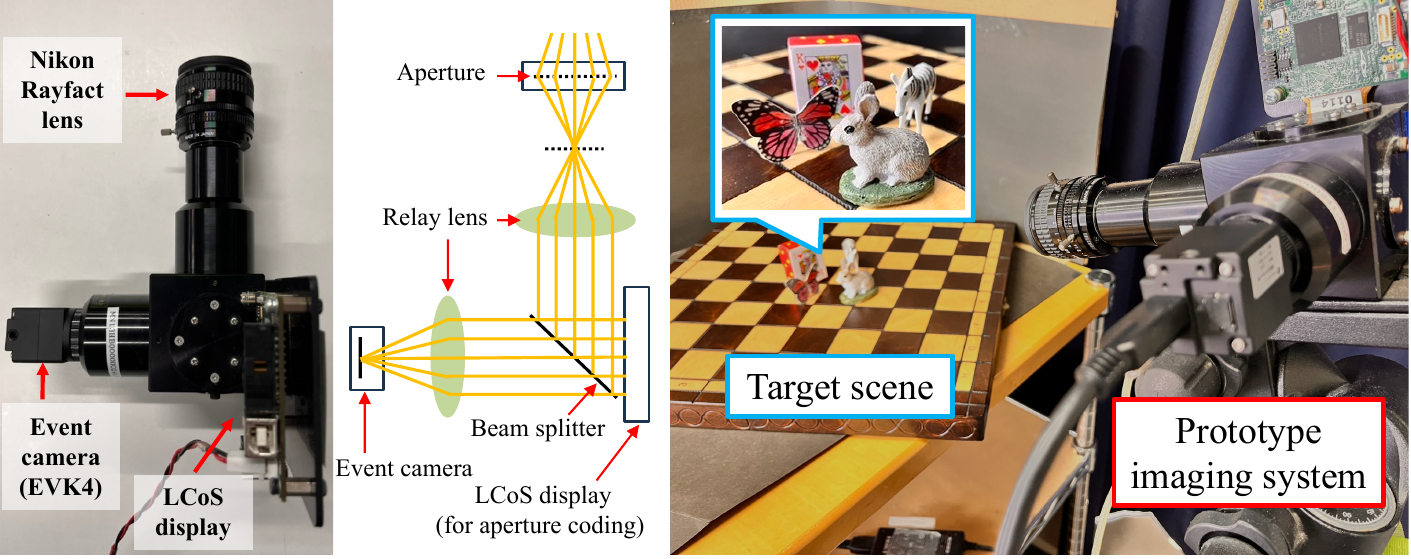} &
\includegraphics[width=\lfW]
{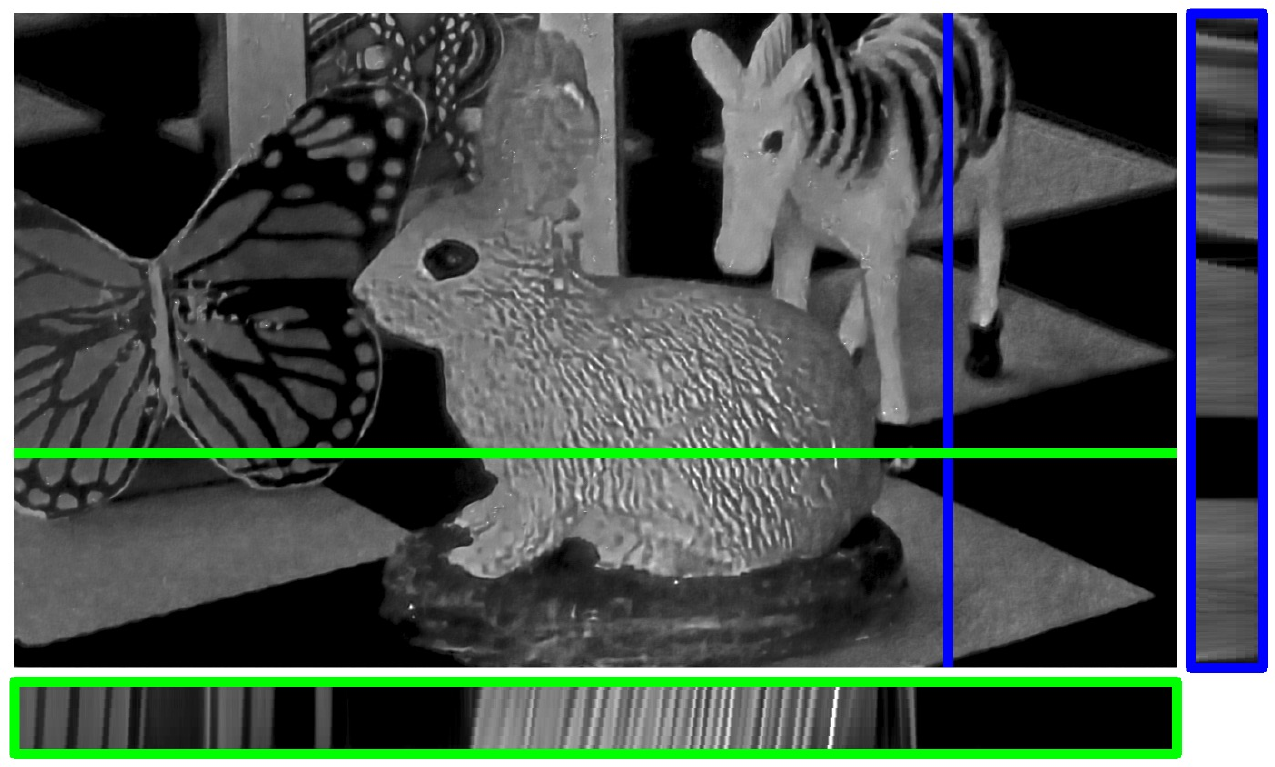} 
\end{tabular}
\caption{Real camera experiment: (left) prototype imaging system, (center) experimental setup, and (right) bottom-right view of reconstructed light field with epipolar plane images (brightness corrected for visualization). See Fig.~\ref{fig:overview} for coding patterns and event images used with this experiment.}
\label{fig:real-exp}
\end{figure*}

\subsection{Real Camera Experiments}

We constructed a prototype imaging system (see Fig.~\ref{fig:real-exp} (left)) to capture real 3-D scenes. The optical system includes a Nikon Rayfact lens (25~mm, F1.4), set of relay optics, beam splitter, and LCoS display (Forth Dimension Displays, SXGA-3DM, $1280 \times 1024 $ pixels). We used the central area of the display ($800 \times 800$ pixels) as the effective aperture and divided it into $8 \times 8$ regions (each with $100 \times 100$ pixels) to display the coding patterns. We used a CenturyArks SilkyEvCam HD camera (event-only, equivalent to an EVK4) to acquire events with $1280\times 720$ pixels. The acquired events were fed to the learned RecNet for light-field reconstruction.

We now give details on the coded aperture. The four coding patterns, $a^{(1)}$, ... , $a^{(4)}$, taken from the learned Baseline+BF+RA were displayed in sequence. Due to hardware restrictions, both a pattern $a^{(n)}$ and its inverse $\bar{a}^{(n)}$ need to be displayed within 50 msec. To meet this restriction, we actually displayed eight patterns in the order of $\bar{a}^{(1)}, \bar{a}^{(2)}, \bar{a}^{(3)}, \bar{a}^{(4)}$, $a^{(1)}, a^{(2)}, a^{(3)}, a^{(4)}$ and used only the events recorded in the duration of $a^{(2)}, a^{(3)}, a^{(4)}$. The duration of $a^{(n)}$ was used as $T^{(n-1,n)}$ in Eq.~(\ref{eq:EventStack}). We observed that the burst of events triggered by a pattern change sometimes continued for a little less than 10 ms; therefore, the time durations of $a^{(n)}$s were set to 10 ms. As an exception, we assigned 15 ms to $a^{(1)}$ to ensure the event stream had settled before the change from $a^{(1)}$ to $a^{(2)}$. Therefore, a single cycle of the coding patterns took 90 ms (94.738 ms including the overhead, which corresponded to 10.6 fps); however, the event measurement for each cycle was conducted in 30 ms (31.302 ms including the overhead).

Figure~\ref{fig:real-exp} also shows the experimental setup (center) and the light field reconstructed from a real 3-D scene (right). Due to the effect of the coded aperture, both the pixel-level intensities and parallax information were embedded in the event images; see Fig.~\ref{fig:overview}. Although not perfect, the reconstructed light field was of convincing visual quality with natural parallax effects; note that this result was achieved with events (three event images) alone. 

In the supplementary video, we present a result with slowly moving objects that were placed on a motorized table rotating at a speed of 1.75 rpm. We recorded events during dozens of cycles of the eight coding patterns and reconstructed a light field for each cycle. Although our method was designed under the static-scene assumption, it was robust to small amounts of motion thanks to the moderate measurement time (approximately 30 ms). This robustness would be improved with faster hardware in the future. 

We also include our software in the supplementary material for demonstrating light field reconstruction from event images captured with our imaging system.

%% file: sec/05_conc.tex
\section{Conclusions}

We proposed Coded-E2LF (coded event to light field), a computational imaging method for acquiring a 4-D light field using a coded aperture and a stationary event-only camera. We introduced theories and methods that enabled pixel-level accurate light-field reconstruction from events alone. In particular, we clarified the key role of a black pattern in the aperture coding patterns. Both simulation and real-camera experiments demonstrated the effectiveness of our method. We believe our study will not only lead to progress in light-field imaging but also broaden the horizon for event-based sensing. 

Our future work includes several directions. We will extend the training dataset and seek better architectures for the reconstruction network to further improve the quality of light field reconstruction. Pursuing a better hardware implementation for a coded aperture is also an interesting avenue. Moreover, we will explore how our method can be extended to capture fast moving scenes; we expect that both hardware-side (a shorter measurement time with faster hardware) and software-side (a motion-aware reconstruction algorithm) approaches will be effective.

%% file: sec/07_supple.tex

\appendix

\section{Additional Results and Discussion}

\subsection{Coding patterns}
\label{sec:supple1}

\begin{figure}[t]
\newcommand{\hs}{\hspace{-0.5mm}}
\newcommand{\myvs}{\vspace{1mm}}
\centering
    \small
    \begin{tabular}{c|cccc}
      & $a^{(1)}$ & $a^{(2)}$ & $a^{(3)}$ & $a^{(4)}$ \\\hline\\[-3mm]
      \raisebox{5mm}{Sequence 0} & 
      \hs\includegraphics[width=10mm,height=10mm]{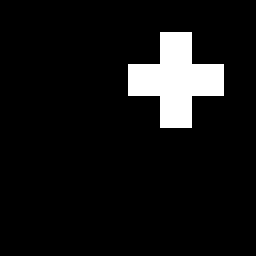}\hs\hs\hs\hs &
      \hs\includegraphics[width=10mm,height=10mm]{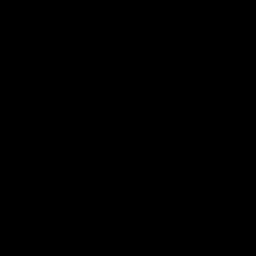}\hs\hs\hs\hs & 
      \hs\includegraphics[width=10mm,height=10mm]{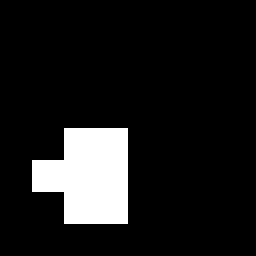}\hs\hs\hs\hs &  
      \hs\includegraphics[width=10mm,height=10mm]{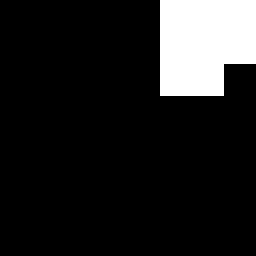}\hs\hs\hs\hs \\\hline\\[-3mm]
      \raisebox{5mm}{Sequence 1} & 
      \hs\includegraphics[width=10mm,height=10mm]{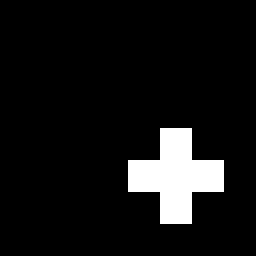}\hs\hs\hs\hs &
      \hs\includegraphics[width=10mm,height=10mm]{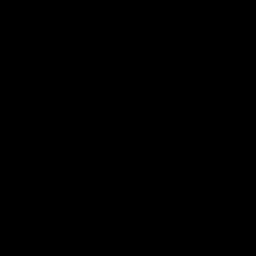}\hs\hs\hs\hs & 
      \hs\includegraphics[width=10mm,height=10mm]{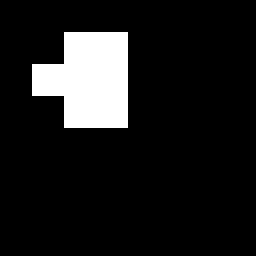}\hs\hs\hs\hs &  
      \hs\includegraphics[width=10mm,height=10mm]{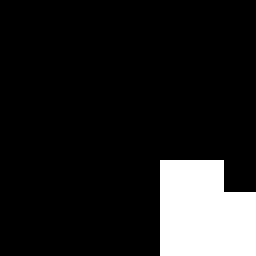}\hs\hs\hs\hs \\\hline\\[-3mm]
      \raisebox{5mm}{Sequence 2} & 
      \hs\includegraphics[width=10mm,height=10mm]{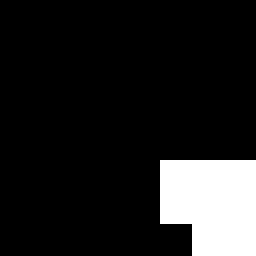}\hs\hs\hs\hs &
      \hs\includegraphics[width=10mm,height=10mm]{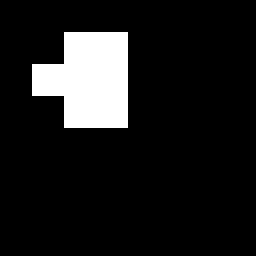}\hs\hs\hs\hs & 
      \hs\includegraphics[width=10mm,height=10mm]{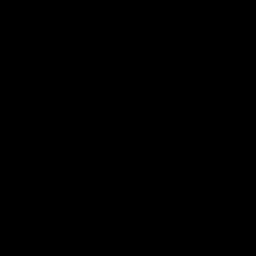}\hs\hs\hs\hs &  
      \hs\includegraphics[width=10mm,height=10mm]{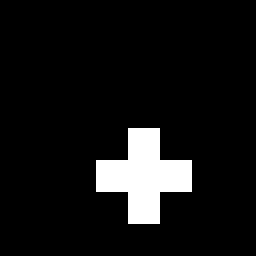}\hs\hs\hs\hs \\\hline\\[-3mm]
      \raisebox{5mm}{Sequence 3} & 
      \hs\includegraphics[width=10mm,height=10mm]{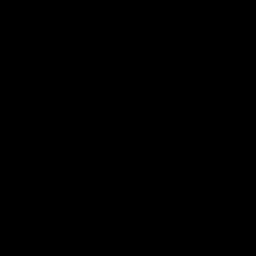}\hs\hs\hs\hs &
      \hs\includegraphics[width=10mm,height=10mm]{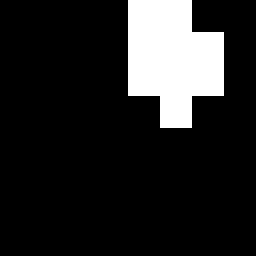}\hs\hs\hs\hs & 
      \hs\includegraphics[width=10mm,height=10mm]{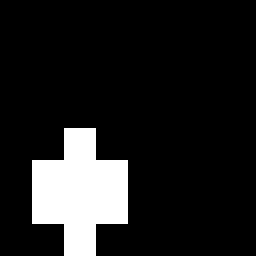}\hs\hs\hs\hs &  
      \hs\includegraphics[width=10mm,height=10mm]{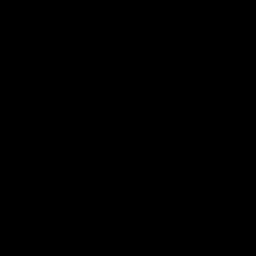}\hs\hs\hs\hs \\\hline\\[-3mm]
      \raisebox{5mm}{Sequence 4} & 
      \hs\includegraphics[width=10mm,height=10mm]{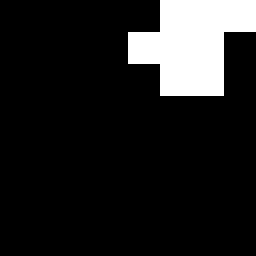}\hs\hs\hs\hs &
      \hs\includegraphics[width=10mm,height=10mm]{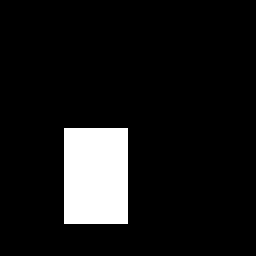}\hs\hs\hs\hs & 
      \hs\includegraphics[width=10mm,height=10mm]{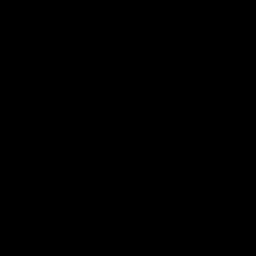}\hs\hs\hs\hs &  
      \hs\includegraphics[width=10mm,height=10mm]{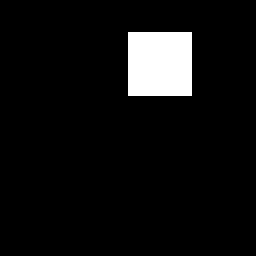}\hs\hs\hs\hs \\\hline\\[-3mm]
      \raisebox{5mm}{Sequence 5} & 
      \hs\includegraphics[width=10mm,height=10mm]{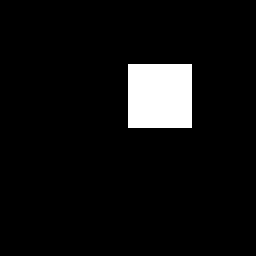}\hs\hs\hs\hs &
      \hs\includegraphics[width=10mm,height=10mm]{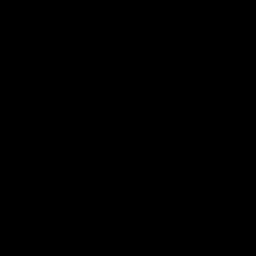}\hs\hs\hs\hs & 
      \hs\includegraphics[width=10mm,height=10mm]{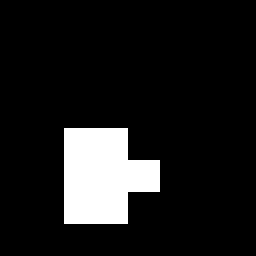}\hs\hs\hs\hs &  
      \hs\includegraphics[width=10mm,height=10mm]{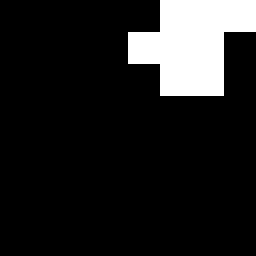}\hs\hs\hs\hs \\\hline\\[-3mm]
      \raisebox{5mm}{Sequence 6} & 
      \hs\includegraphics[width=10mm,height=10mm]{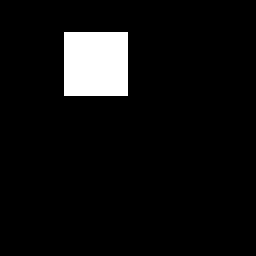}\hs\hs\hs\hs &
      \hs\includegraphics[width=10mm,height=10mm]{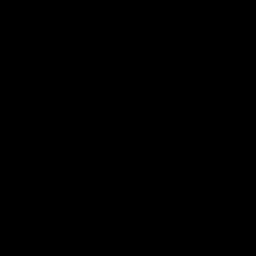}\hs\hs\hs\hs & 
      \hs\includegraphics[width=10mm,height=10mm]{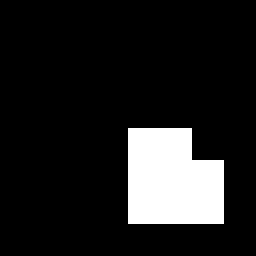}\hs\hs\hs\hs &  
      \hs\includegraphics[width=10mm,height=10mm]{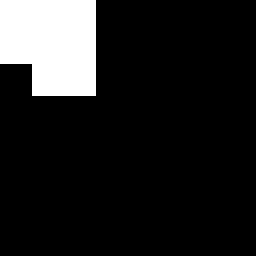}\hs\hs\hs\hs \\\hline\\[-3mm]
      \raisebox{5mm}{Sequence 7} & 
      \hs\includegraphics[width=10mm,height=10mm]{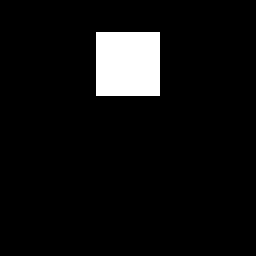}\hs\hs\hs\hs &
      \hs\includegraphics[width=10mm,height=10mm]{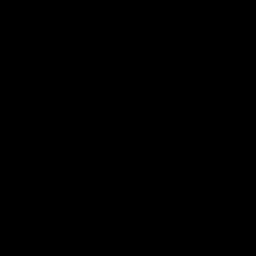}\hs\hs\hs\hs & 
      \hs\includegraphics[width=10mm,height=10mm]{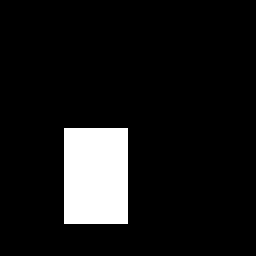}\hs\hs\hs\hs &  
      \hs\includegraphics[width=10mm,height=10mm]{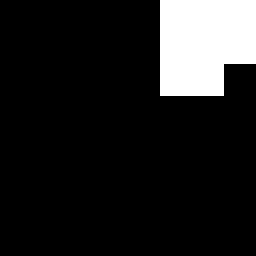}\hs\hs\hs\hs \\\hline\\[-3mm]
      \raisebox{5mm}{Sequence 8} & 
      \hs\includegraphics[width=10mm,height=10mm]{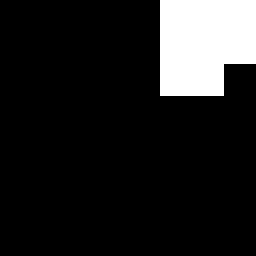}\hs\hs\hs\hs &
      \hs\includegraphics[width=10mm,height=10mm]{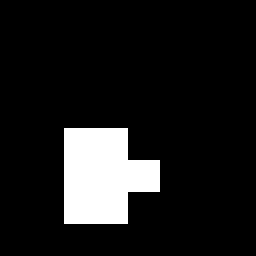}\hs\hs\hs\hs & 
      \hs\includegraphics[width=10mm,height=10mm]{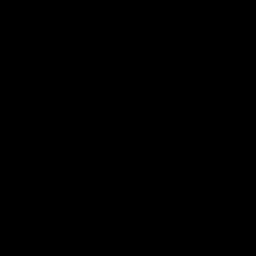}\hs\hs\hs\hs &  
      \hs\includegraphics[width=10mm,height=10mm]{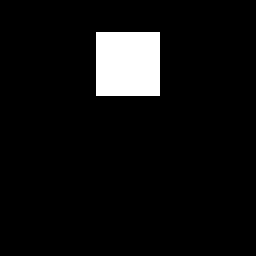}\hs\hs\hs\hs \\\hline\\[-3mm]
      \raisebox{5mm}{Sequence 9} & 
      \hs\includegraphics[width=10mm,height=10mm]{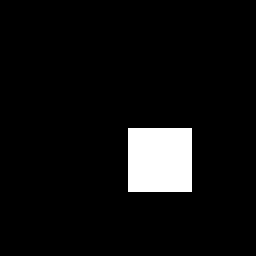}\hs\hs\hs\hs &
      \hs\includegraphics[width=10mm,height=10mm]{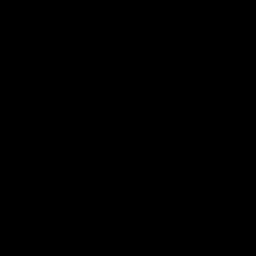}\hs\hs\hs\hs & 
      \hs\includegraphics[width=10mm,height=10mm]{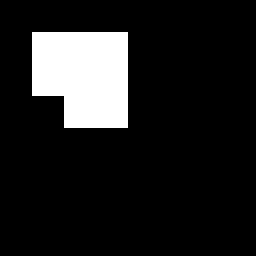}\hs\hs\hs\hs &  
      \hs\includegraphics[width=10mm,height=10mm]{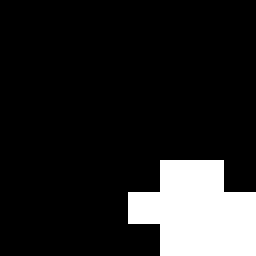}\hs\hs\hs\hs \\\hline\\[-3mm]
      \raisebox{5mm}{Sequence 10} & 
      \hs\includegraphics[width=10mm,height=10mm]{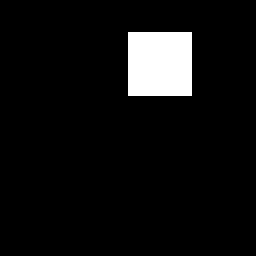}\hs\hs\hs\hs &
      \hs\includegraphics[width=10mm,height=10mm]{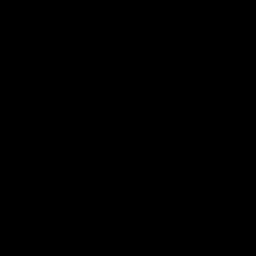}\hs\hs\hs\hs & 
      \hs\includegraphics[width=10mm,height=10mm]{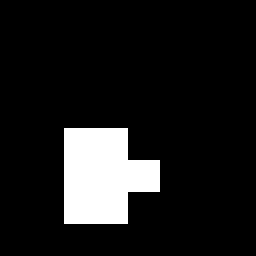}\hs\hs\hs\hs &  
      \hs\includegraphics[width=10mm,height=10mm]{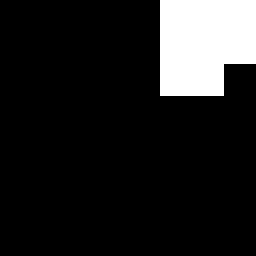}\hs\hs\hs\hs \\

    \end{tabular}
\caption{Coding patterns learned with baseline algorithm.}
\label{fig:pattern-examples}
\end{figure}

 In Fig.~\ref{fig:pattern-examples}, we present examples of coding patterns learned with our baseline algorithm. Note that at least one black pattern was included in each coding-pattern sequence.

The coding patterns have limited light efficiency, as the number of white elements is less than ten. This phenomenon results from a trade-off between the signal-to-noise ratio (SNR) and defocus blur; higher brightness leads to better SNR but increases blurring. 

\subsection{Additional ablation study}
\label{sec:supple2}

\begin{figure}[t]
\newcommand{\hs}{\hspace{-1mm}}
\newcommand{\myvs}{\vspace{1mm}}
\newcommand{\HS}{\hspace{2mm}}
\newcommand{\BS}{\hspace{-13mm}}
\newcommand{\bs}{\hspace{-11mm}}
\centering
    \small
    \begin{tabular}{c|cccc}
      Method & \HS $a^{(1)}$ & \HS $a^{(2)}$ & \HS $a^{(3)}$ & \HS $a^{(4)}$ \\
      (total events) &  & \BS $|E^{(1,2)}|$ &  \BS $|E^{(2,3)}|$ &  \BS $|E^{(3,4)}|$ \\ \hline\\[-3mm]
      
      \raisebox{5mm}{Baseline+RA} & 
      \hs\includegraphics[width=12mm,height=12mm]{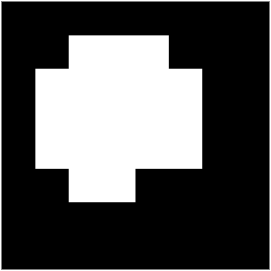}\hs\hs\hs\hs &
      \hs\includegraphics[width=12mm,height=12mm]{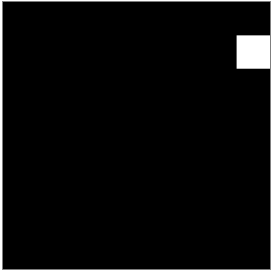}\hs\hs\hs\hs & 
      \hs\includegraphics[width=12mm,height=12mm]{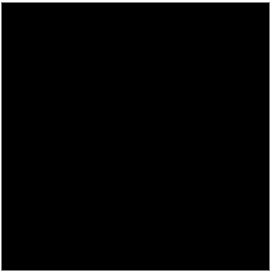}\hs\hs\hs\hs &  
      \hs\includegraphics[width=12mm,height=12mm]{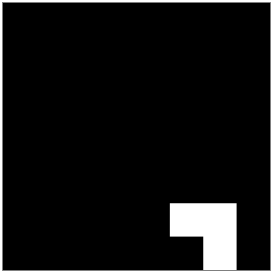}\hs\hs\hs\hs \\
      (10.785) &  & \bs 6.413 & \bs 1.704 & \bs 2.668 \\ \hline \\[-3mm]      
    \end{tabular}
    \caption{Learned coding patterns ($N=4$) with event counts in $E^{(1,2)}$, $E^{(2,3)}$, and $E^{(3,4)}$, and their total. Event counts are per pixel, averaged in BasicLFSR test dataset.}
    \label{fig:baseline+ra}
\end{figure}

In the main paper, we report three variants of our method (Baseline, Baseline+BF, Baseline+BF+RA). Here, we additionally report the result with Baseline+RA, where we initialize $I^{\mathrm{ref}}_{x,y} = I^{(1)}_{x,y}$ for all $(x,y)$s at $n=1$ and use Eqs.~(\ref{eq:ESIM2}) and (\ref{eq:ESIM2-update}) for the event generation and reference update processes. The learned coding patterns are presented in Fig.~\ref{fig:baseline+ra}. As shown in Table~\ref{tab:ablation}, Baseline and Baseline+RA were almost equivalent in terms of the reconstruction quality. The best reconstruction quality was achieved with Baseline+BF+RA, which is the full implementation of our method.

The difference in the total event counts between Baseline+BF (6.522) and Baseline+BF+RA (7.175) is mainly attributed to the difference in the event generation models. Baseline+BF was trained and evaluated using the baseline event generation model (Eq.~(\ref{eq:ESIM})), which underestimated the number of events. Meanwhile, Baseline+BF+RA was trained and evaluated using the reference-aware event generation model (Eqs.~(\ref{eq:ESIM2}) and (\ref{eq:ESIM2-update})). If the coding patterns of Baseline+BF was evaluated using Eqs.~(\ref{eq:ESIM2}) and (\ref{eq:ESIM2-update}), the total event count increased to 7.273 events/pixel.

\begin{table}[t]
\caption{Summary of ablation study with $N=4$. All reported scores are averaged in BasicLFSR test dataset. }
\label{tab:ablation}
\begin{tabular}{c||c|c}
Method & Total event count & PSNR/SSIM \\
\hline
Baseline & 9.492 &  29.81/0.8302\\
Baseline+RA & 10.785 & 29.87/0.8309\\
Baseline+BF & 6.522 &  29.69/0.8321\\
\textbf{Baseline+BF+RA} & 7.175 &  30.48/0.8413\\
\end{tabular}

\end{table}

\subsection{Comparison with intensity-based methods}
\label{sec:supple3}

For simulation experiments, we strictly followed the protocols of Habuchi et al.~\cite{Habuchi_2024_CVPR}; thus, our results can be directly compared with those in their paper~\cite{Habuchi_2024_CVPR}. Some results with intensity-based methods are summarized in Table~\ref{tab:compare} to show the position of our method. Intensity-based coded aperture imaging (CA + RecNet) offers a clear advantage in terms of the reconstruction fidelity. However, our method has the potential to reduce the measurement time for light fields even in low light conditions.

\begin{table}[t]
\caption{Comparison with intensity-based methods. CA: coded aperture, LA: lens-array, @I: number of intensity images, @E: number of event images. $*$ indicates quotation from Habuchi et al.'s~\cite{Habuchi_2024_CVPR}. All six methods share the same RecNet architecture as Habuchi et al.'s~\cite{Habuchi_2024_CVPR}. }
\label{tab:compare}
\vspace{-2.5mm}
\centering
{\small
\begin{tabular}{c|c|c|c}
Methods & @I & @E & PSNR[dB]/SSIM (ALL) \\
\hline\hline
CA + RecNet* & 4 & -- & 35.39/0.9346 \\
CA + RecNet* & 2 & -- & 34.09/0.9210\\
CA + RecNet* & 1 & -- & 27.62/0.8139\\
LA + RecNet* & 1 & --  & 26.22/0.7218\\
\hline
Habuchi~\cite{Habuchi_2024_CVPR}  & 1 & 3 & 31.83/0.8729\\
Ours  & -- & 3  & 30.48/0.8413\\
\end{tabular}
}
\end{table}

\subsection{Data rate analysis}
\label{sec:supple4}

\begin{figure}[t]
    \centering
    \includegraphics[width=\linewidth]{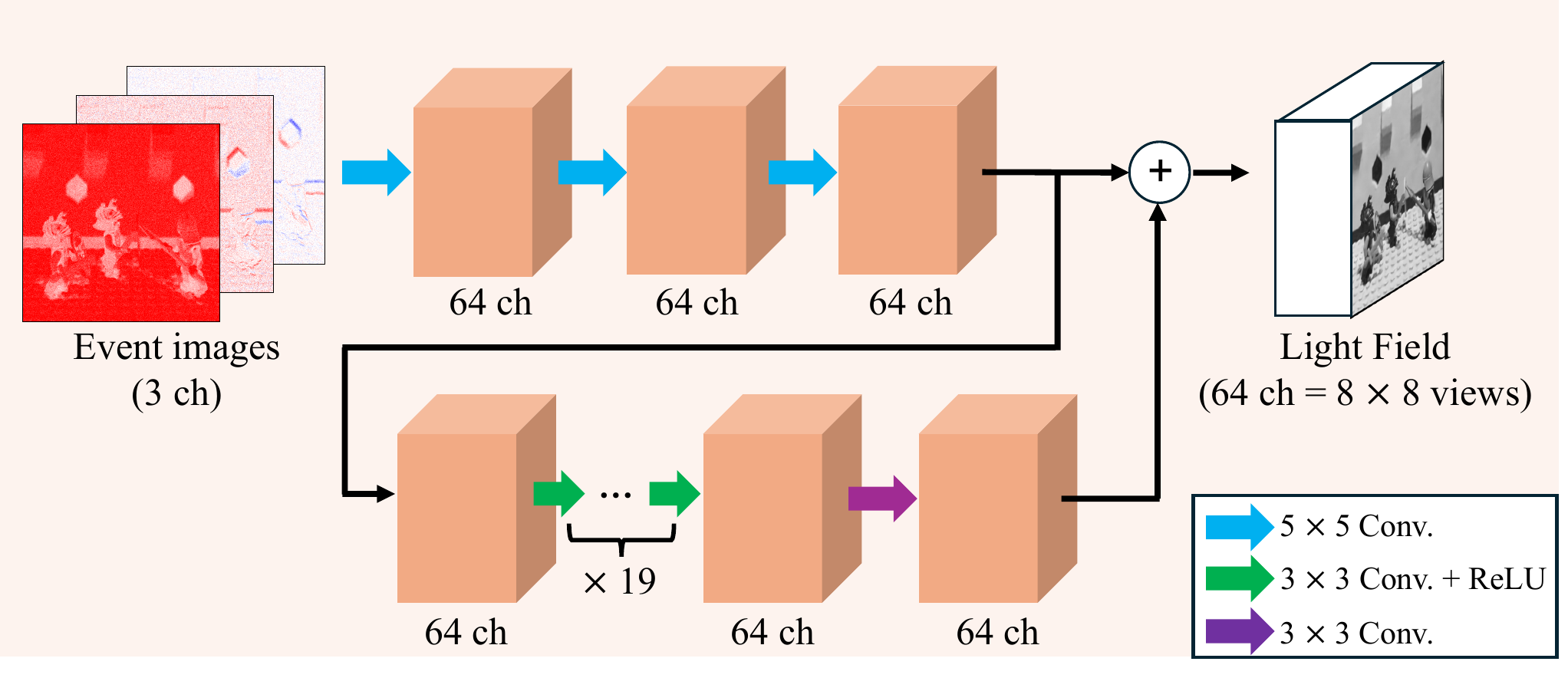}
    \caption{Network architecture of RecNet}
    \label{fig:recnet}
\end{figure}

Unfortunately, it is difficult to claim the advantage of our method in terms of data rate or bandwidth compared to its intensity-based counterpart. Since the event count depends on the brightness/complexity of the target scene, we discuss the average on the test dataset. Our method captures 7.175 events/pixel at the sensor, which corresponds to 0.112 events/pixel with respect to all the pixels of a light field ($8 \times 8$ views). If each event requires 29 bits in the COO format, the data rate is 208 bits/pixel at the sensor and 3.25 bits/pixel with respect to light field pixels. This is more efficient than direct acquisition of $8 \times 8$ views, but less efficient than the intensity-based coded aperture method (e.g., 8 bits/pixel $\times$ 2 -- 4 frames).

\section{Network architecture of RecNet}
\label{sec:supple-recnet}

For RecNet, we used the same architecture as Habuchi et al.~\cite{Habuchi_2024_CVPR}, which is illustrated in Fig.~\ref{fig:recnet}.